\documentclass[runningheads]{llncs}
\usepackage[T1]{fontenc}
\usepackage{graphicx}
\usepackage{amsmath}
\usepackage{amssymb}
\usepackage{subcaption}
\usepackage{booktabs}
\usepackage{multirow}

\usepackage{hyperref}
\usepackage{color}

\begin{document}

\setlength{\intextsep}{6pt plus 2pt minus 2pt}
\setlength{\floatsep}{6pt plus 2pt minus 2pt}
\setlength{\textfloatsep}{8pt plus 2pt minus 2pt}
\captionsetup[table]{aboveskip=2pt, belowskip=6pt}

\title{On the Efficiency of LoRA Fine-Tuning for Vision-Language-Action Models in Industrial Robotic Manipulation}
\titlerunning{Tuning VLA Models for Industrial Robotic Manipulation}

\author{Finn Ferchau\inst{1,2} \and Daniel Pommer\inst{1}\orcidID{0009-0002-4109-4763} \and
\\ Cristian Axenie\inst{1,3}\orcidID{0000-0001-6184-0546}}

\authorrunning{F. Ferchau, D. Pommer, and C. Axenie}

\institute{Technische Hochschule Nürnberg Georg Simon Ohm, Nürnberg, Germany \\
\email{{cristian.axenie, daniel.pommer}@th-nuernberg.de} \and
Siemens AG, München, Germany \\ 
\email{finn.ferchau@siemens.com} \and
Fraunhofer Institute for Integrated Circuits (IIS), Erlangen, Germany
\email{cristian.axenie@iis.fraunhofer.de}
}

\maketitle

\begin{abstract}
Deploying billion-parameter Vision-Language-Action (VLA) models on industrial hardware requires fine-tuning to bridge the embodiment gap. Full Fine-Tuning (FFT) provides maximal plasticity but requires data centre-grade GPUs. We present a systematic study of Low-Rank Adaptation (LoRA) for $\pi_0$, a flow-matching VLA, evaluated on four precision assembly tasks with a UR5e robotic manipulator. Across a sweep of LoRA ranks ($r{=}8$ to $256$), allocation strategies, and component-freezing ablations, we find no statistically significant advantage of FFT over certain LoRA configurations. Performance saturates at $r{=}32$, and uniform allocation across the Vision-Language-Model (VLM) backbone and action expert proves sufficient. Freezing the VLM or restricting the vision encoder to LoRA significantly degrades performance, indicating that embodiment adaptation requires both semantic and visual plasticity. These results suggest that LoRA at $r{=}32$ with full vision encoder fine-tuning is a practical approach, reducing static peak VRAM from 36.2 to 10.8\,GiB (parameters and optimizer states, activation memory excluded) without detectable performance loss.
\end{abstract}

\section{Introduction}

Industrial assembly has traditionally relied on precisely programmed robotic systems that excel in repetitive, high-volume production but require costly reprogramming when tasks or parts change~\cite{hagele2016industrial}. Vision-Language-Action (VLA) models promise to address this rigidity by enabling robots to execute manipulation tasks from natural-language instructions, thereby replacing explicit programming with learning from demonstrations~\cite{rt22023arxiv,black2024pi0}. However, deploying these foundation models on specific industrial hardware is not straightforward: a critical \emph{embodiment gap} exists between the heterogeneous pre-training data and the kinematics of a target manipulator, necessitating fine-tuning for real-world deployment~\cite{black2024pi0}.

Full Fine-Tuning (FFT) updates all model parameters and provides maximal plasticity, but for large models, it requires data centre-grade GPUs. This often forces practitioners to transfer proprietary training data to external cloud providers, potentially conflicting with industrial data privacy requirements~\cite{shi2016edge}. Low-Rank Adaptation (LoRA)~\cite{hu2021lora} offers a parameter-efficient alternative by injecting trainable low-rank matrices while keeping the pre-trained weights frozen, reducing GPU memory requirements.

While LoRA has shown promising results in natural language processing and vision tasks, its application to VLA models for robotic manipulation raises two questions that prior work has not addressed. First, \emph{how does robotic manipulation performance scale with LoRA rank}---the factorization dimension that bounds the true matrix rank of the weight update? Existing evaluations test one or two configurations without characterizing the full capacity--performance curve. Second, and specific to flow-matching VLAs: \emph{where should adapter capacity be allocated?} Flow-matching VLAs such as $\pi_0$~\cite{black2024pi0} separate a VLM backbone from a dedicated action expert (AE) that generates continuous motor commands (Section~\ref{sec:related_vla}). Yet, no prior work has investigated how to distribute a fixed adapter budget between these two components.

We address both questions through a systematic experimental study on a UR5e manipulator. Our contributions span five axes:

\begin{enumerate}
    \item \textbf{Rank scaling.} We evaluate LoRA-based fine-tuning at different ranks (e.g. at $r{=}8, 16, 32, 64, 128, 256$), bounded by a zero-shot and an FFT baseline.
    \item \textbf{Parameter allocation.} To test whether adapter capacity is better spent on the VLM or the action expert, we compare two asymmetric configurations: ($r_{\text{VLM}}{=}16$, $r_{\text{AE}}{=}128$) and ($r_{\text{VLM}}{=}128$, $r_{\text{AE}}{=}16$).
    \item \textbf{VLM necessity.} We freeze the VLM entirely and apply FFT to the action expert alone, testing whether embodiment adaptation is exclusively a motor control problem.
    \item \textbf{Vision encoder adaptation.} We vary the treatment of the Sigmoid Loss for Language Image Pre-training (SigLIP) vision encoder---full fine-tuning, LoRA, or frozen---to determine whether LoRA's performance depends on unconstrained visual adaptation.
    \item \textbf{Intrinsic rank analysis.} We perform singular value decomposition (SVD) of the FFT weight deltas to characterize the intrinsic dimensionality of embodiment adaptation.
\end{enumerate}

\noindent All models are trained and evaluated on the same task suite with identical data, and we document the VRAM requirements of each. We additionally release open-source hardware integrations and training code to facilitate reproducibility.\footnote{\url{https://github.com/F-Fer/lerobot_ur5e_gello}, \url{https://github.com/F-Fer/openpi-ur5e}, video demonstrations: \url{https://www.youtube.com/playlist?list=PLVRwcikxp4E9UU1L6PlqYIp3nvq5rf0Yp}}

\section{Related Work}

\subsection{Vision-Language-Action Models}\label{sec:related_vla}
Learning-based approaches to robotic manipulation have progressed from behaviour cloning with compact policy networks~\cite{zhao2023act} to billion-parameter foundation models that fuse visual perception, natural language understanding, and motor control within a single architecture. RT-1~\cite{brohan2023rt1} introduced a Transformer-based policy mapping image sequences and language instructions to tokenized actions, and its successor RT-2~\cite{rt22023arxiv} formalized the VLA paradigm by co-training on web-scale vision--language data alongside robot demonstrations. Subsequent open-source efforts broadened access: OpenVLA~\cite{kim2024openvla} provided an open foundation model trained on the Open X-Embodiment corpus~\cite{oxe2023}, while Octo~\cite{octo2024} explored diffusion-based conditioning on language and goal images. More recent work, such as NeuroVLA\cite{guo2026brain}, which is based on StarVLA (StarVLA: A Lego-like Codebase for Vision-Language-Action Model Developing \cite{starvla2025}), also explicitly exploits temporal sparsity to minimise end-to-end latency, enabling localised, hardware-efficient learning on edge devices.

A key architectural split has emerged in how VLAs generate actions. \emph{Autoregressive} models such as OpenVLA and $\pi_0$-FAST~\cite{pertsch2025fast} discretise actions into tokens produced sequentially, which simplifies training but limits the frequency of control. \emph{Flow-matching} models, exemplified by $\pi_0$, instead predict an entire \emph{action chunk} in parallel via a learned vector field, yielding smooth, high-frequency actions. $\pi_0$ pairs an adapted PaliGemma VLM backbone~\cite{beyer2024paligemma} with a dedicated \emph{action expert} that performs the flow-matching denoising.

More recent models such as GR00T~\cite{gr00tn1_2025} and SmolVLA~\cite{shukor2025smolvla} adopt related dual-system or compact flow-matching designs, while $\pi_{0.5}$~\cite{black2025pi05} extends the paradigm with hierarchical co-training for open-world generalization.

\subsection{Parameter-Efficient Fine-Tuning for VLAs}
LoRA freezes a pre-trained weight matrix $W_0 \in \mathbb{R}^{d \times k}$ and injects a low-rank update $\Delta W = BA$, where $B \in \mathbb{R}^{d \times r}$ and $A \in \mathbb{R}^{r \times k}$ with $r \ll \min(d,k)$. The modified forward pass becomes $h = W_0 x + \frac{\alpha}{r}\,BAx$, where $h$ is the layer output, $x$ the input, and $\alpha$ a scaling factor. Only $A$ and $B$ are optimized, reducing both memory consumption and trainable parameter count.

The OpenVLA study~\cite{kim2024openvla} compared adaptation strategies and reported that LoRA matched FFT performance while updating only 1.4\,\% of parameters. Since their evaluation was limited to two LoRA configurations, it is open how performance scales with rank and how adapter capacity should be distributed.

\section{Experiment Setup}
\label{sec:setup}

\subsection{Hardware Platform}
\begin{figure}[h]
    \centering
    \includegraphics[width=0.7\linewidth]{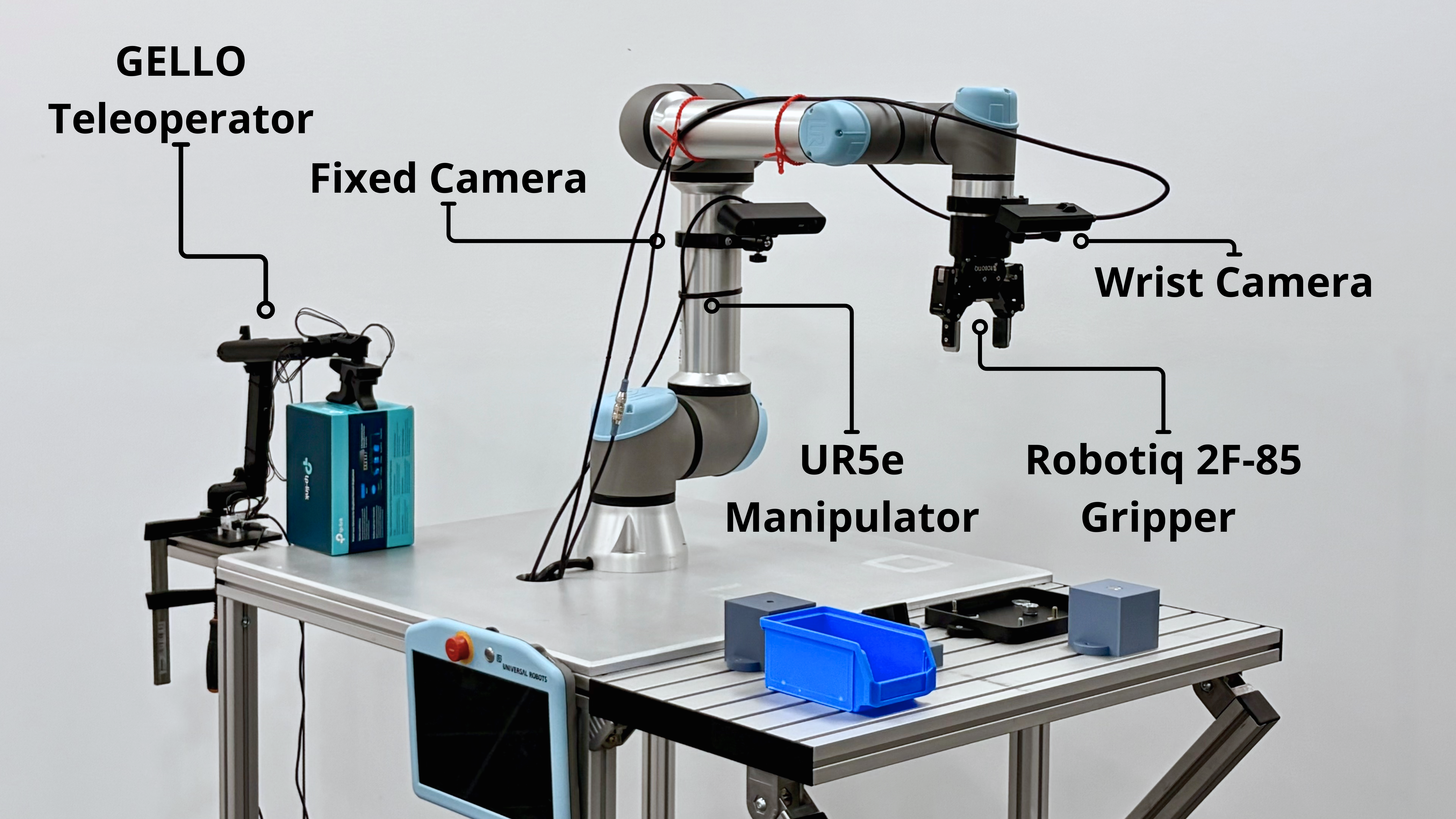}
    \caption{Experimental Hardware Setup. UR5e robotic platform with GELLO teleoperator and stereo camera configuration.}
    \label{fig:hw_setup}
\end{figure}

All experiments are conducted on a Universal Robots UR5e, a 6 degree of freedom (DoF) manipulator equipped with a Robotiq 2F-85 gripper. Visual observations are captured by two Stereolabs cameras: a ZED~2i mounted on the robot's upper arm link provides a global workspace view, while a wrist-mounted ZED~Mini supplies a stereo egocentric perspective. Together, the three resulting RGB streams form the observation space consumed by the $\pi_0$ policy. Expert demonstrations are collected with a GELLO teleoperation device~\cite{wu2023gello}. This 3D-printed leader arm is kinematically isomorphic to the UR5e and maps joint-space commands directly to the follower at 60\,Hz (see Fig.~\ref{fig:hw_setup}).\footnote{Video demonstrations of the hardware setup and task executions are available at \url{https://youtube.com/playlist?list=PLEhuTFwREAwg1jdXL1mme3czLyfDI6pA5&si=wZJxeckM3v8iZ5OD}.}

\noindent To enable this setup within the $\pi_0$ ecosystem, we extended the open-source LeRobot~\cite{cadene2024lerobot} and OpenPi~\cite{OpenPiGitHub} frameworks with custom hardware plugins for the UR5e, GELLO, and a ZeroMQ-based camera streaming pipeline.

\subsection{Task Suite}

\begin{figure}[h]
    \centering
    \begin{subfigure}[b]{0.24\textwidth}
        \includegraphics[width=\linewidth]{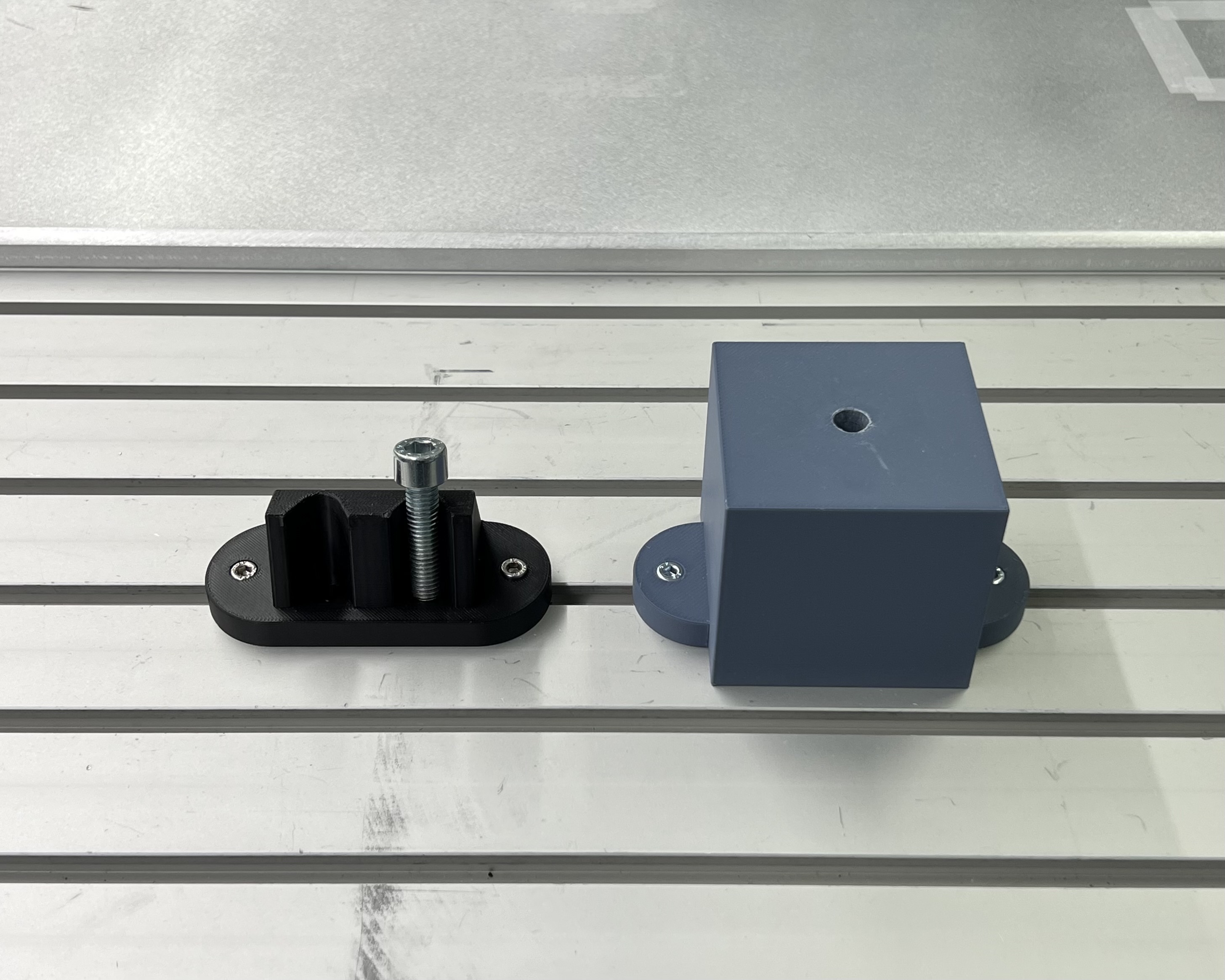}
    \end{subfigure}
    \hfill
    \begin{subfigure}[b]{0.24\textwidth}
        \includegraphics[width=\linewidth]{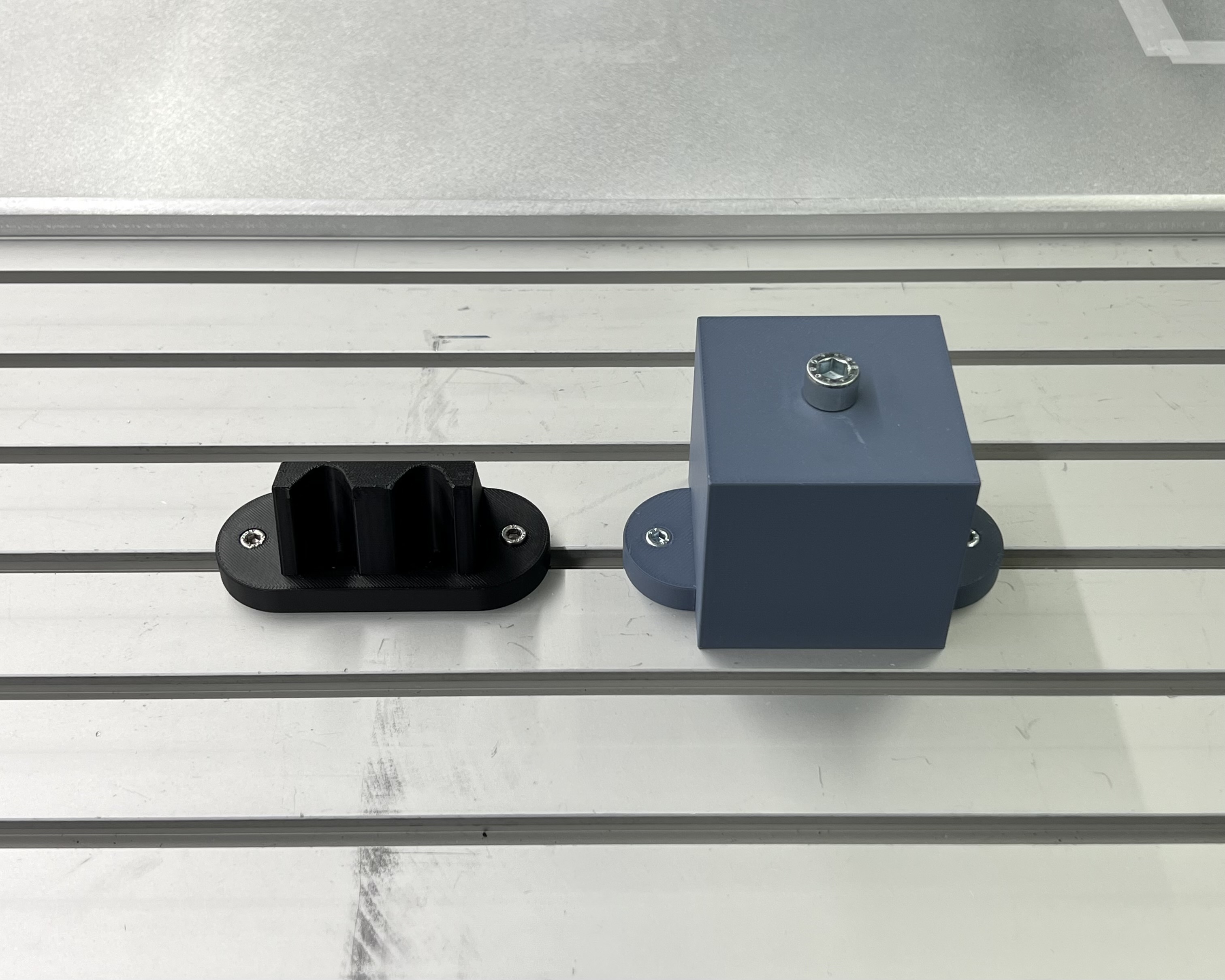}
    \end{subfigure}
    \hfill
    \begin{subfigure}[b]{0.24\textwidth}
        \includegraphics[width=\linewidth]{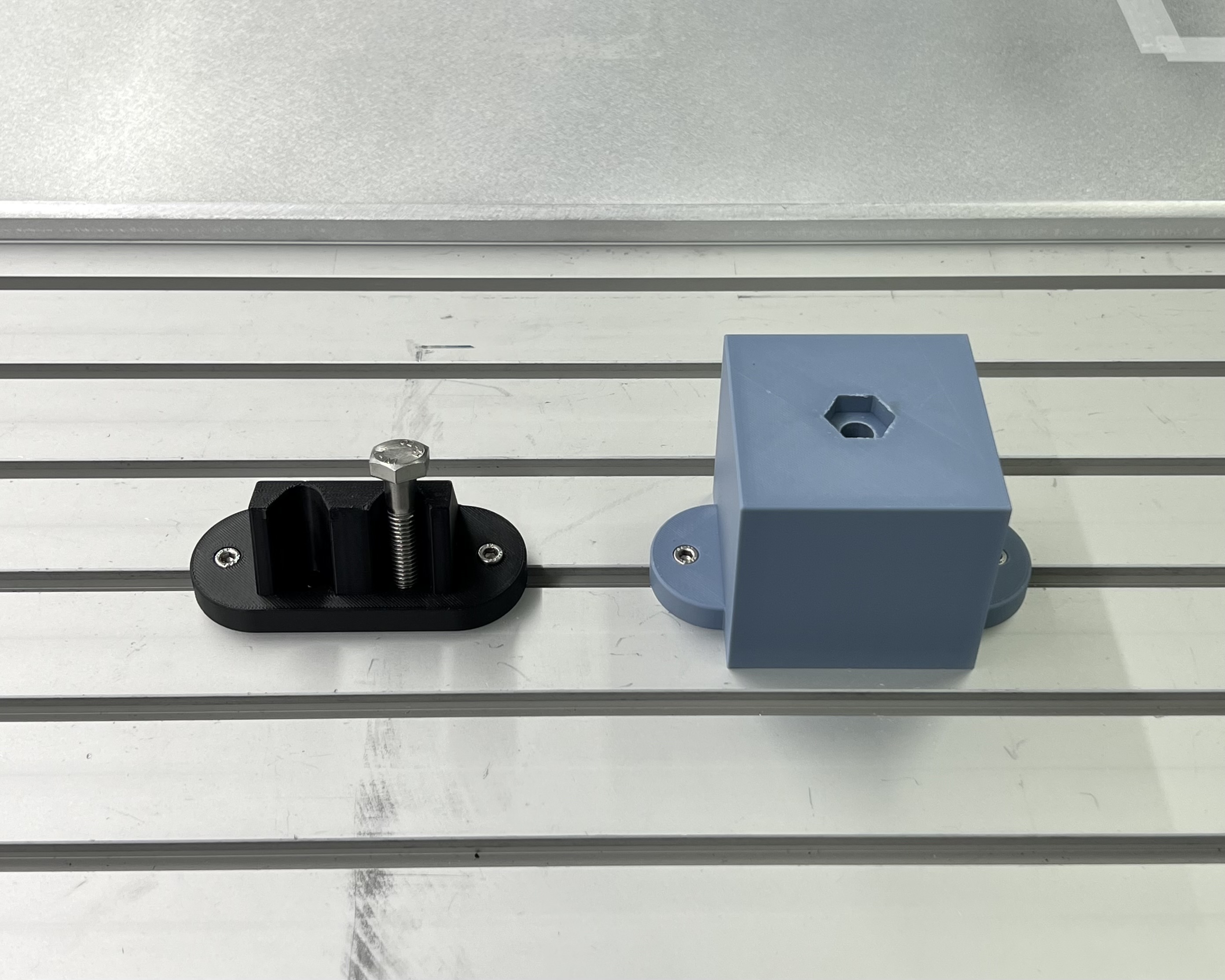}
    \end{subfigure}
    \hfill
    \begin{subfigure}[b]{0.24\textwidth}
        \includegraphics[width=\linewidth]{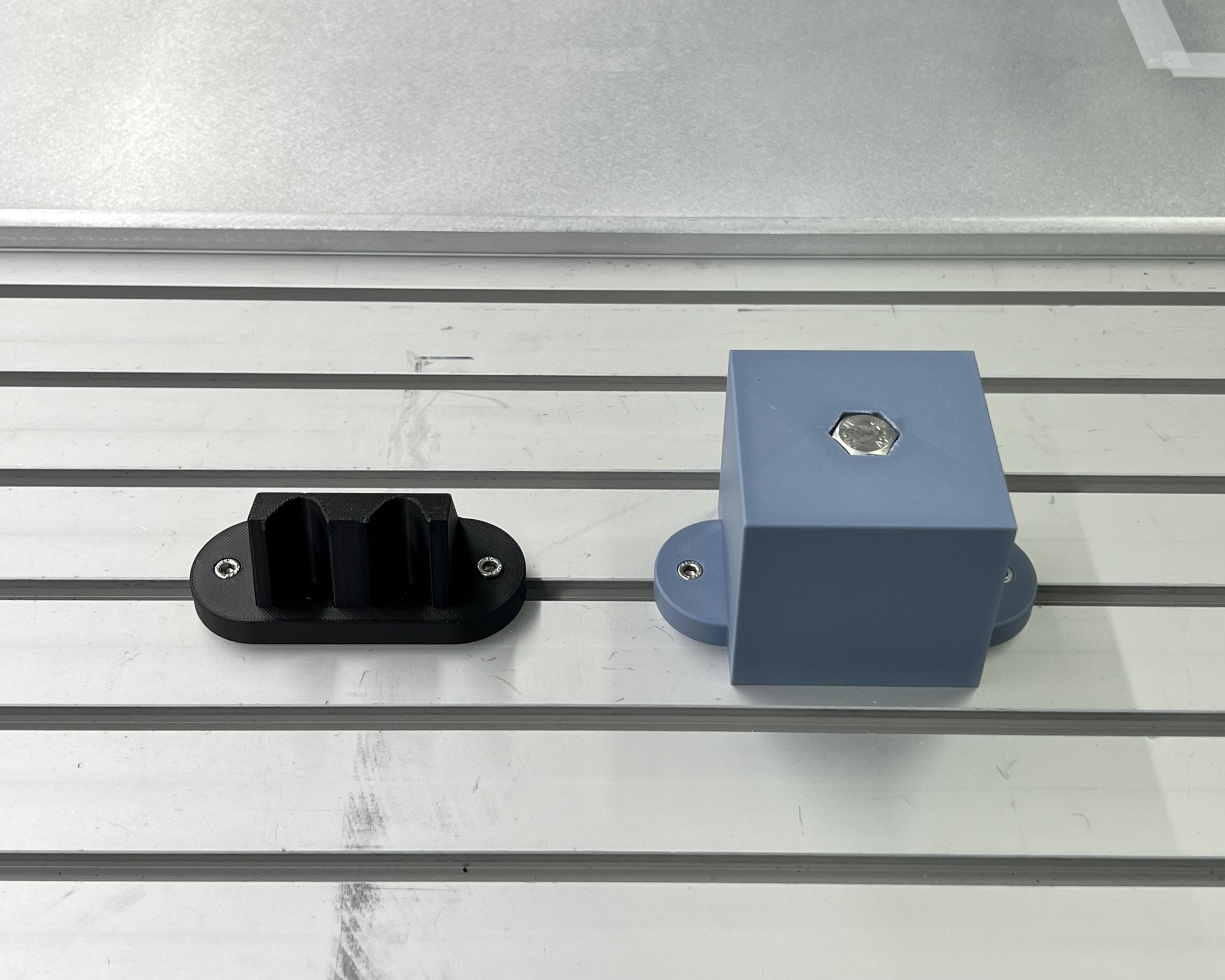}
    \end{subfigure}

    \vspace{0.4em}

    \begin{subfigure}[b]{0.24\textwidth}
        \includegraphics[width=\linewidth]{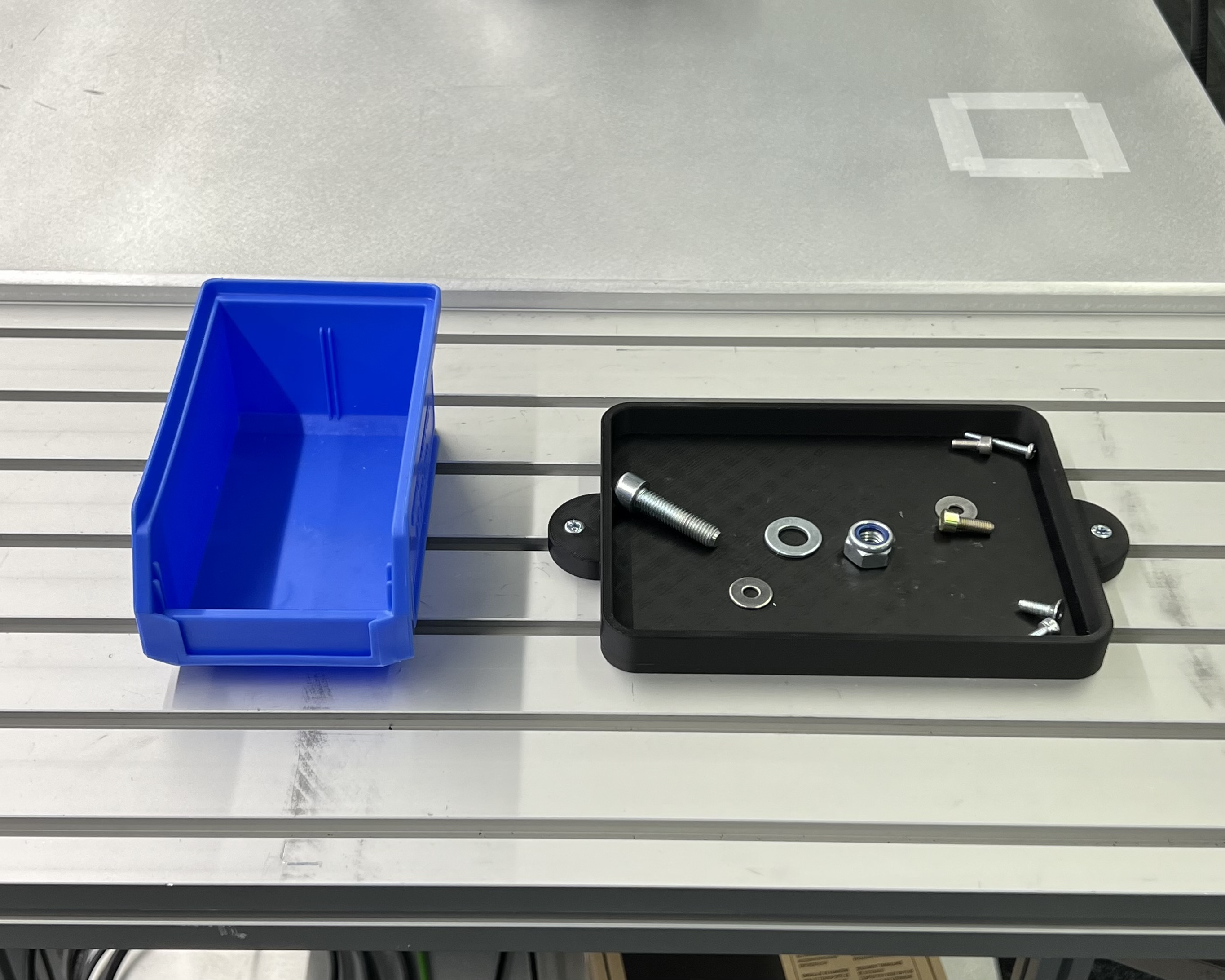}
    \end{subfigure}
    \hfill
    \begin{subfigure}[b]{0.24\textwidth}
        \includegraphics[width=\linewidth]{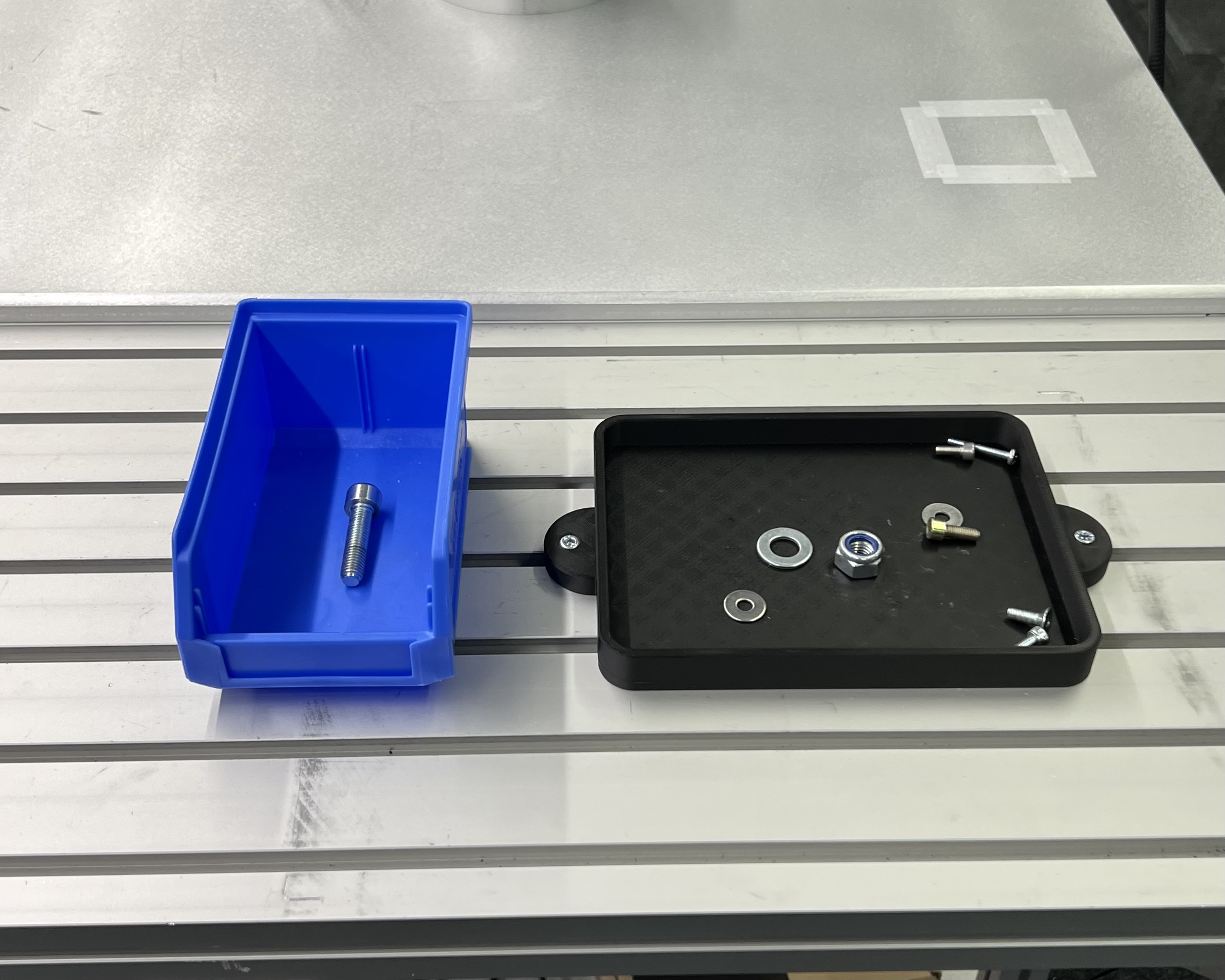}
    \end{subfigure}
    \hfill
    \begin{subfigure}[b]{0.24\textwidth}
        \includegraphics[width=\linewidth]{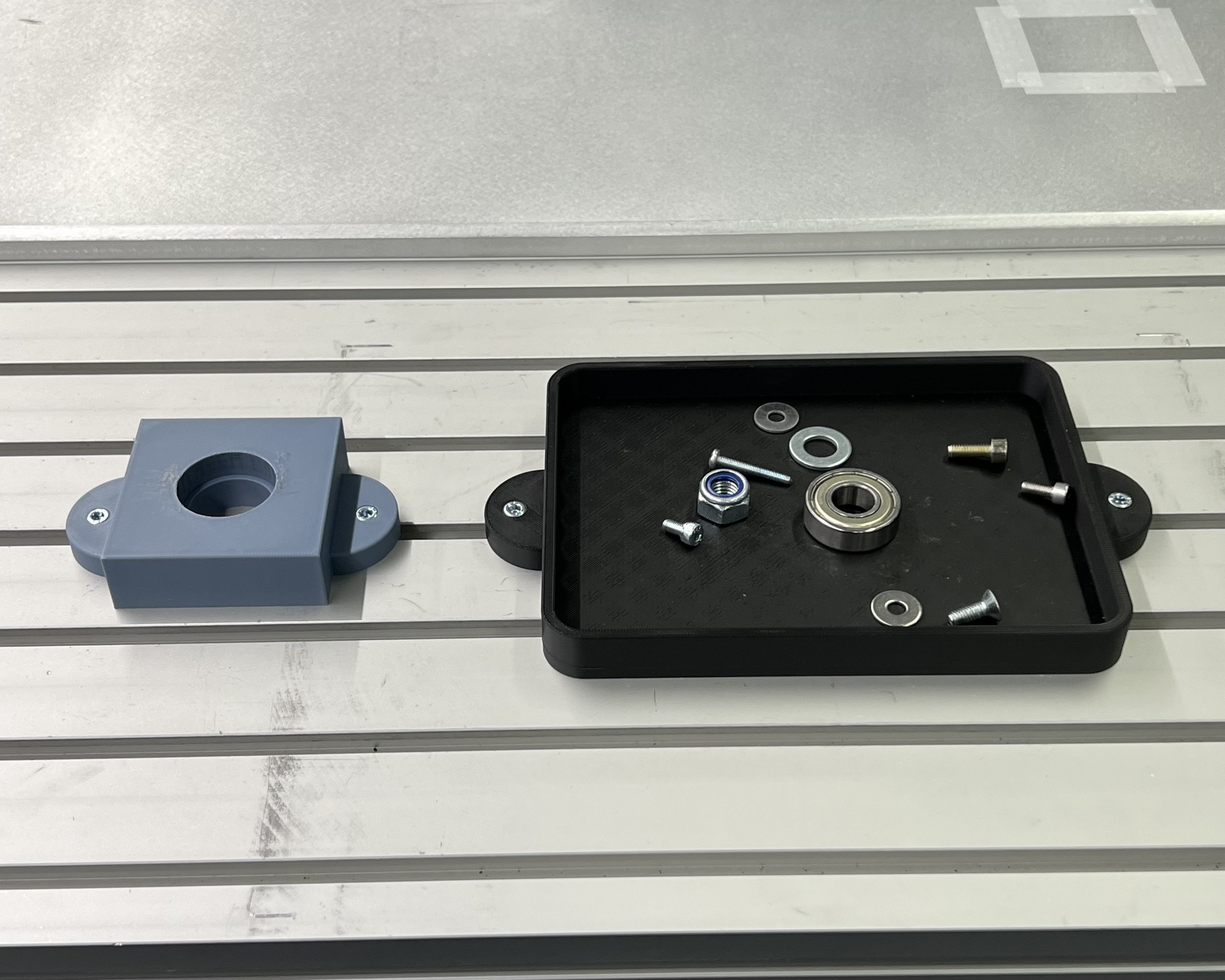}
    \end{subfigure}
    \hfill
    \begin{subfigure}[b]{0.24\textwidth}
        \includegraphics[width=\linewidth]{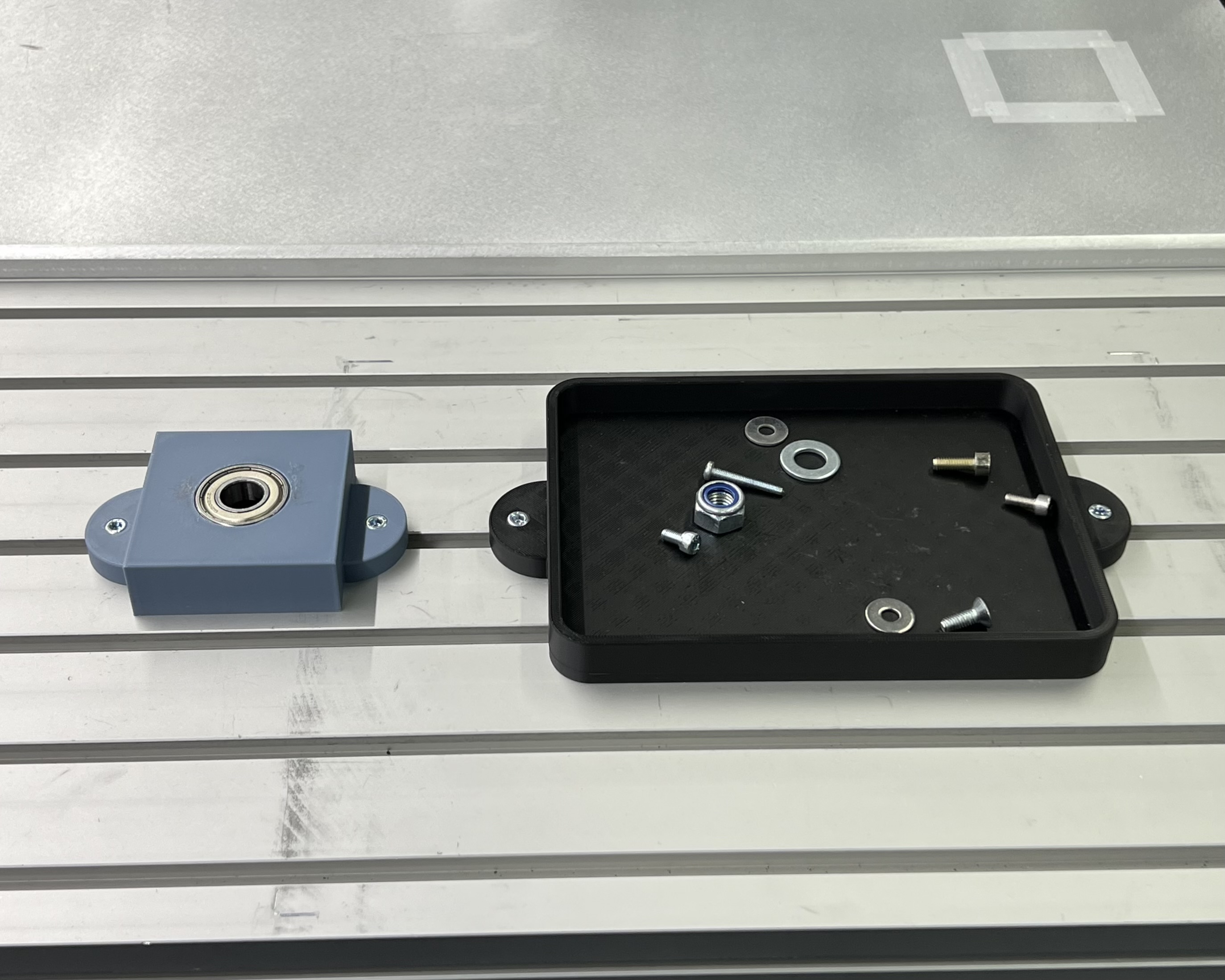}
    \end{subfigure}

    \caption{Task suite overview showing initial (left) and goal (right) states. Top: Bolt Insertion Easy / Hard. Bottom: Pick \& Place / Bearing Press-Fit.}
    \label{fig:tasks}
\end{figure}

\noindent We evaluate on four assembly tasks that span a range of industrial manipulation challenges (Fig.~\ref{fig:tasks}).

\noindent The four tasks are:

\begin{enumerate}
    \item \textbf{Bolt Insertion (Easy):} Grasp a bolt from a fixture and insert it into a cylindrical hole. Requires position accuracy in $(x, y, z)$.
    \item \textbf{Bolt Insertion (Hard):} Insert a bolt into a hexagonal recess, requiring joint position and orientation alignment.
    \item \textbf{Pick \& Place:} Identify the correct bolt among distractors on a parts tray and place it into a bin.
    \item \textbf{Bearing Press-Fit:} Grasp a bearing and press it flush into a housing. A contact-rich task requiring sustained downward force against friction.
\end{enumerate}

\noindent Each task is decomposed into sequential sub-goals used to compute the evaluation metric described in Section~\ref{sec:eval}.

\subsection{Data Collection}
For each task, 200 expert demonstrations were collected, yielding 800 episodes totalling approximately 5.5 hours of synchronized data. Each episode records joint positions (6-DoF), gripper state (1-DoF), the teleoperator-demanded manipulator and gripper position (action, 7-DoF), and three RGB streams at $672{\times}376$ resolution, all at 60\,Hz. The episodes are stored in the LeRobot Dataset v3.0 format. Before training, per-dimension mean and standard deviation are computed over the full 800-episode corpus for the state and action vectors; these statistics are used for normalization during training and inference.

\subsection{Training Configurations}
\label{sec:training_config}
Table~\ref{tab:configs} summarizes the evaluated configurations.

\begin{table}[h]
    \centering
    \caption{Overview of evaluated configurations. All LoRA runs use
    $\alpha{=}r$.}
    \label{tab:configs}
    \begin{tabular}{l c c r}
        \hline
        \textbf{Configuration} & $r_{\text{VLM}}$ & $r_{\text{AE}}$
            & \textbf{Trainable Params} \\
        \hline
        Zero-shot                  & -- & -- & 0 \\
        LoRA $r{=}8$               & 8   & 8   & 438\,M \\
        LoRA $r{=}16$              & 16  & 16  & 457\,M \\
        LoRA $r{=}32$              & 32  & 32  & 496\,M \\
        LoRA $r{=}64$              & 64  & 64  & 574\,M \\
        LoRA $r{=}128$             & 128 & 128 & 729\,M \\
        LoRA $r{=}256$             & 256 & 256 & 1,041\,M \\
        LoRA AE-heavy              & 16  & 128 & 534\,M \\
        LoRA VLM-heavy             & 128 & 16  & 652\,M \\
        AE-only FFT (VLM frozen)   & --  & all & 315\,M \\
        LoRA $r{=}32$ + SigLIP LoRA   & 32 & 32 & 91\,M \\
        LoRA $r{=}32$ + SigLIP frozen & 32 & 32 & 81\,M \\
        FFT                        & all & all & 3\,238\,M \\
        \hline
    \end{tabular}
\end{table}

\noindent All models are trained with the AdamW optimizer, a cosine learning rate schedule (1\,000 warmup steps, peak $2.5{\times}10^{-5}$, decay to $2.5{\times}10^{-6}$), and an action horizon of $H{=}30$ steps. LoRA adapters target the query, key, value, and output projections, as well as the feedforward layers, in both the VLM and the action expert. The SigLIP vision encoder remains fully fine-tuned across all LoRA configurations unless otherwise specified. Two additional configurations vary the SigLIP treatment to test whether visual adaptation requires full plasticity. Following Hu et al.~\cite{hu2021lora}, we set $\alpha{=}r$ so that the effective scaling factor $\alpha/r{=}1$ remains constant across ranks. We adopt this convention so that rank is the only variable that changes across the sweep, isolating its effect on task performance under the standard LoRA protocol.

FFT and LoRA $r{\in}\{128, 256\}$ experiments are conducted on a single NVIDIA H100 (80\,GB) with batch sizes of 32 and 24, respectively. All remaining experiments run on a single NVIDIA RTX 4090 (24\,GB) with batch size 24. All configurations use mixed-precision training (bfloat16 activations; float32 optimizer states for FFT, bfloat16 backbone weights for LoRA).

\subsection{Intrinsic Rank Analysis}
\label{sec:intrinsic_rank_method}
To understand why low-rank adaptation can match FFT, we analyze the intrinsic dimensionality of FFT weight updates via SVD. For each adapted layer, we compute the weight delta $\Delta W = W_{\text{fine-tuned}} - W_{\text{pre-trained}}$ and decompose it as $\Delta W = U \Sigma V^\top$. We restrict the analysis to multi layer perceptron (MLP) layers (gating and down-projection weights), which account for the majority of the parameters in each component. We analyze all three adapted components---VLM, action expert, and SigLIP vision encoder---each comprising 18, 18, and 27 transformer layers, respectively. As a summary metric, we report the rank $k$ at which the cumulative squared singular values reach a fraction $\tau$ of the total Frobenius norm: $\sum_{i=1}^{k} \sigma_i^2 \geq \tau \cdot \|\Delta W\|_F^2$, with $\tau{=}0.95$ as the default threshold.

\section{Evaluation Methodology}
\label{sec:eval}

Each configuration from Table~\ref{tab:configs} is evaluated with 20 physical rollouts per task.
Since binary success rates can obscure meaningful distinctions, we follow recent VLA evaluations~\cite{black2024pi0,shukor2025smolvla,gr00tn1_2025} and adopt Average Task Progress (ATP). Each task is decomposed into $S_{\text{total}}$ sequential sub-goals, weighted equally. ATP is defined as:
\begin{equation}
    \text{ATP} = \frac{1}{N} \sum_{i=1}^{N} \frac{S^{(i)}_{\text{completed}}}{S_{\text{total}}}
    \label{eq:atp}
\end{equation}
where $S^{(i)}_{\text{completed}}$ is the number of sub-goals completed in rollout $i$ and $N{=}20$. This scoring increases statistical power at moderate sample sizes of 10--20 trials per task~\cite{black2024pi0,kim2024openvla}.

\section{Results}
\label{sec:results}

Our analysis and experimental results revealed five key findings that will be made available to the community when considering the practical deployment of VLA models.

\paragraph{Finding 1: LoRA Shows No Statistically Significant Performance Drop.}

\begin{figure}[h]
    \centering
    \includegraphics[width=0.5\linewidth]{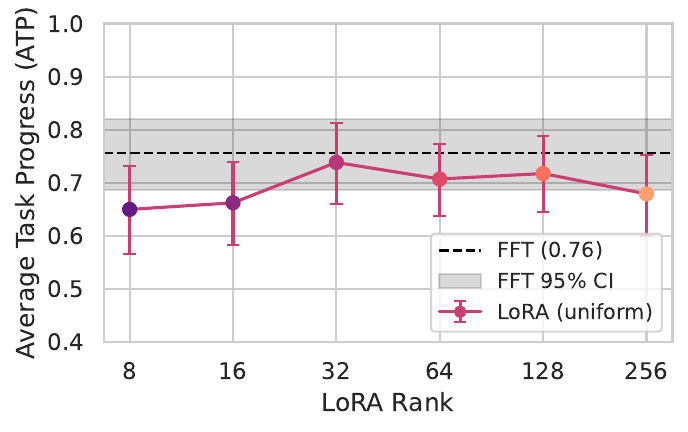}
    \caption{Average Task Progress as a function of LoRA rank ($r{=}8$ to $256$). The dashed line and shaded band indicate FFT mean and 95\,\% bootstrap CI.}
    \label{fig:rank_scaling}
\end{figure}

As shown in Table~\ref{tab:atp_overall}, uniform LoRA configurations with a rank $r>32$ show performance that falls into the 95\% interval of the FFT reference and their confidence intervals overlapped substantially with it, suggesting no meaningful difference in performance under this evaluation protocol.
The best LoRA variant ($r{=}32$) falls within two percentage points of FFT, with a negligible rank-biserial effect size.
Mann-Whitney $U$ tests with Holm-Bonferroni correction found no significant differences between any LoRA configuration and FFT (all $p > 0.05$). In contrast, Freeze VLM and both SigLIP-restricted configurations are significantly worse ($p < 0.001$). The related Figure~\ref{fig:rank_scaling} shows that ATP performance increases from $r{=}8$ to $r{=}32$ and then plateaus. This implies that a rank of 32 captures the task-relevant adaptation.

\begin{table}[h]
    \centering
    \caption{Overall ATP at 60k steps with Mann-Whitney $U$ tests against FFT (Holm-Bonferroni corrected). $\pm$: half-width of the 95\,\% bootstrap confidence Interval (CI) (10,000 resamples). $r$: rank-biserial effect size. $^{*}p < 0.001$.}
    \label{tab:atp_overall}
    \setlength{\tabcolsep}{6pt}
    \begin{tabular}{l c c c c}
        \toprule
        \textbf{Configuration} & \textbf{ATP} & \textbf{95\,\% CI} & \textbf{$p_{\text{holm}}$} & \textbf{Effect} ($r$) \\
        \midrule
        FFT                     & \textbf{0.76} & $\pm$0.07 & -- & -- \\
        LoRA $r{=}256$          & 0.68 & $\pm$0.08 & 1.000  & $-$0.107 (small) \\
        LoRA $r{=}128$          & 0.72 & $\pm$0.08 & 1.000  & $-$0.058 (negligible) \\
        LoRA $r{=}64$           & 0.71 & $\pm$0.07 & 1.000  & $-$0.095 (negligible) \\
        LoRA $r{=}32$           & 0.74 & $\pm$0.08 & 1.000  & 0.006 (negligible) \\
        LoRA $r{=}16$           & 0.66 & $\pm$0.08 & 0.808  & $-$0.142 (small) \\
        LoRA $r{=}8$            & 0.65 & $\pm$0.08 & 0.808  & $-$0.141 (small) \\
        LoRA VLM-heavy          & 0.69 & $\pm$0.07 & 0.977  & $-$0.121 (small) \\
        LoRA AE-heavy           & 0.69 & $\pm$0.07 & 1.000  & $-$0.105 (small) \\
        Freeze VLM              & 0.15 & $\pm$0.06 & $<$0.001$^{*}$ & $-$0.804 (large) \\
        LoRA $r{=}32$ + SigLIP LoRA   & 0.43 & $\pm$0.09 & $<$0.001$^{*}$ & $-$0.492 (medium) \\
        LoRA $r{=}32$ + SigLIP frozen & 0.14 & $\pm$0.07 & $<$0.001$^{*}$ & $-$0.764 (large) \\
        Zero-shot               & 0.00 & -- & -- & -- \\
        \bottomrule
    \end{tabular}
\end{table}

\paragraph{Finding 2: Uniform Allocation Is Sufficient.}

Both asymmetric allocations---VLM-heavy and AE-heavy---perform at or below the uniform $r{=}32$ configuration, despite using more trainable parameters. This suggests that both components benefit from adaptation and that concentrating capacity on either one is suboptimal.

\paragraph{Finding 3: VLM and Vision Encoder Adaptation Are Essential.}
\label{sec:siglip_results}

Freezing the VLM and fine-tuning only the action expert yields ATP${=}0.15$, significantly below FFT ($p < 0.001$, large effect; Table~\ref{tab:atp_overall}), indicating that embodiment adaptation is not exclusively a motor control problem. Similarly, freezing the SigLIP vision encoder yields ATP${=}0.14$, comparable to the VLM-frozen result. Applying LoRA to SigLIP partially recovers performance (ATP${=}0.43$) but remains significantly worse than the baseline LoRA $r{=}32$ with SigLIP FFT ($p < 0.001$, medium effect). Together, these results indicate that both semantic and visual plasticity are necessary for effective adaptation.

\begin{figure}[h]
    \centering
    \includegraphics[width=0.85\linewidth]{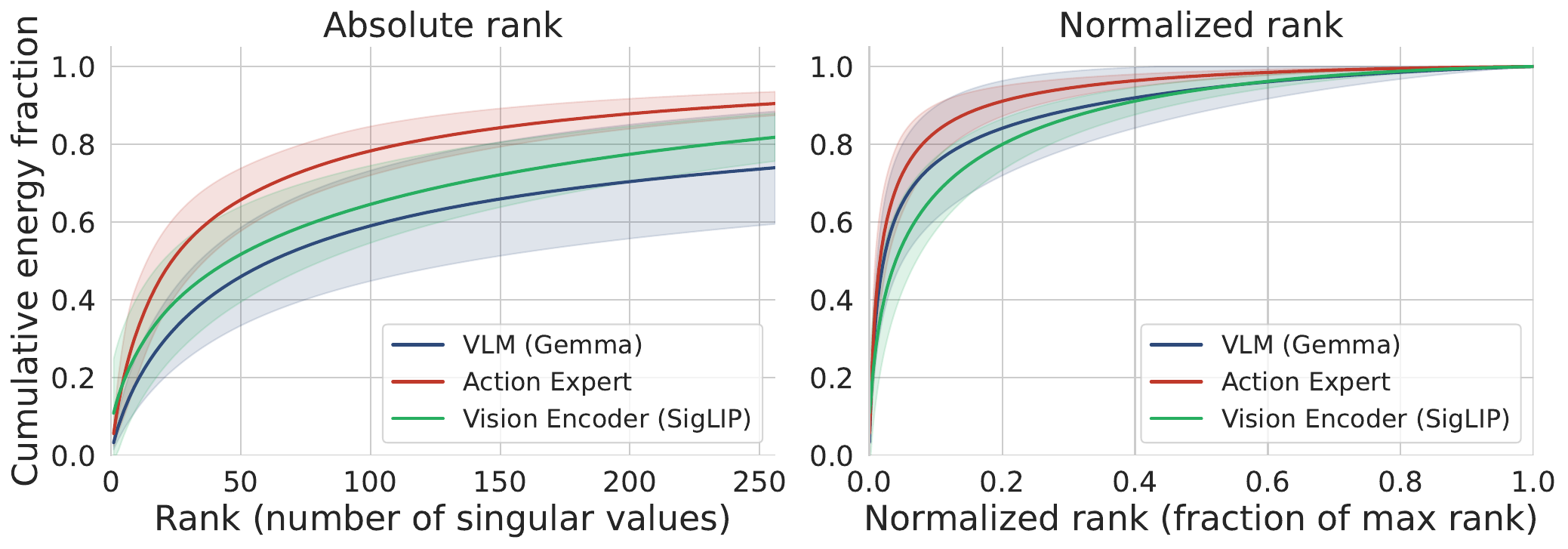}
    \caption{Cumulative spectral energy of FFT weight deltas for MLP layers. Left: absolute rank. Right: rank normalized by each matrix's maximum rank ($\min(m,n)$). Shaded regions indicate ${\pm}1$ standard
 deviation across layers.}
    \label{fig:cumulative_energy}
\end{figure}

\paragraph{Intrinsic Rank of FFT Weight Updates Is a Structural Property.}
\label{sec:intrinsic_rank_results}
Figure~\ref{fig:cumulative_energy} shows, the cumulative energy of FFT weight deltas for MLP layers in each component. The rank required to capture 95\,\% of the Frobenius norm energy is substantial across all components: $441{\pm}133$ for action expert MLPs, $597{\pm}135$ for SigLIP vision encoder MLPs, and $1{,}402{\pm}579$ for the VLM MLPs. At common LoRA ranks, the energy captured is modest---for instance, at $r{=}32$ the action expert MLPs capture ${\sim}57\,\%$ of the spectral energy and the VLM MLPs only ${\sim}37\,\%$.

\noindent After normalizing rank by the maximum possible rank ($\min(m,n)$ of each weight matrix) to control for architectural differences in hidden dimension, the action expert MLPs consistently use a smaller fraction of their available spectrum (${\sim}31\,\%$ at 95\,\% energy) than the VLM MLPs (${\sim}47\,\%$) or the SigLIP vision encoder MLPs (${\sim}52\,\%$). This ordering---visible in the normalized panel of Fig.~\ref{fig:cumulative_energy}---indicates that the lower intrinsic dimensionality of action expert MLP updates is a structural property, not an artifact of its smaller model dimension, and that the SigLIP vision encoder MLPs undergo the most distributed adaptation of all three MLP components.

\begin{figure}[h]
    \centering
    \includegraphics[width=0.75\linewidth]{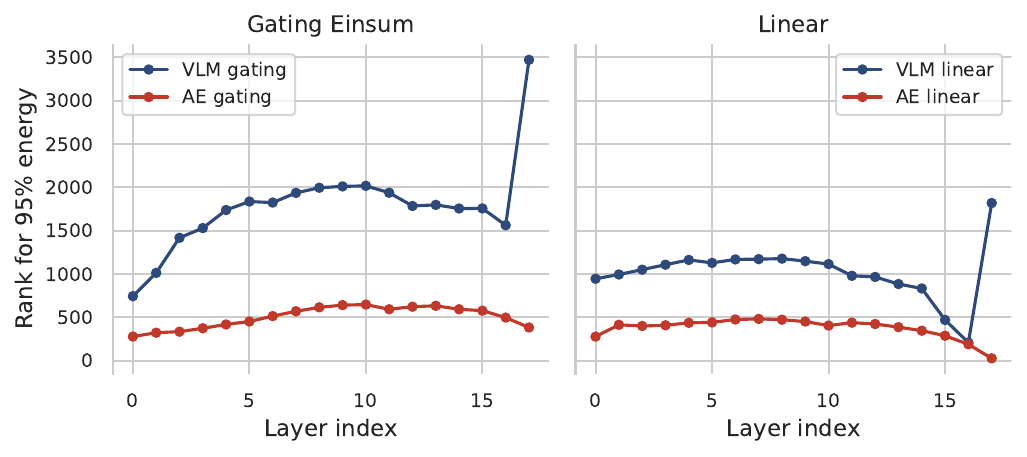}
    \caption{Rank required to capture 95\,\% of spectral energy of FFT weight deltas for MLP layers per transformer layer.}
    \label{fig:intrinsic_rank_per_layer}
\end{figure}

A per-layer breakdown (Fig.~\ref{fig:intrinsic_rank_per_layer}) reveals pronounced heterogeneity in intrinsic rank across transformer layers. The VLM's last MLP layer exhibits a sharp spike---reaching ${\sim}3{,}500$ for gating einsum weights (out of 4{,}096 maximum singular values) and ${\sim}1{,}800$ for linear weights---while the action expert's last layers show a decrease.

\paragraph{Memory and Efficiency.}
\label{sec:memory}

As reported in Table~\ref{tab:memory}, the trainable parameter counts and VRAM usage are different for each evaluated model. LoRA at $r{=}32$ reduces trainable parameters to 15.0\,\% of FFT and static peak VRAM for the model from 36.2\,GiB to an estimated 10.8\,GiB, enabling training on consumer-grade GPUs.

\begin{table}[h]
    \centering
    \caption{Trainable parameters and model VRAM usage for each configuration. VRAM estimates include parameters and optimizer states; activation memory is excluded.}
    \label{tab:memory}
    \begin{tabular}{l r r r}
        \toprule
        \textbf{Configuration} & \textbf{Trainable (M)} & \textbf{Trainable (\%)} & \textbf{VRAM (GiB)} \\
        \midrule
        FFT                     & 3{,}238 & 100.0 & 36.2 \\
        LoRA $r{=}256$          & 1{,}041 & 27.0  & 16.9 \\
        LoRA $r{=}128$          & 729     & 20.6  & 13.4 \\
        LoRA VLM-heavy          & 652     & 18.8  & 12.5 \\
        LoRA $r{=}64$           & 574     & 16.9  & 11.7 \\
        LoRA AE-heavy           & 534     & 15.9  & 11.2 \\
        LoRA $r{=}32$           & 496     & 15.0  & 10.8 \\
        LoRA $r{=}16$           & 457     & 14.0  & 10.4 \\
        LoRA $r{=}8$            & 438     & 13.4  & 10.1 \\
        AE-only FFT             & 315     & 9.7   & 9.0 \\
        LoRA $r{=}32$ + SigLIP LoRA   & 91  & 2.7 & 7.0 \\
        LoRA $r{=}32$ + SigLIP frozen & 81  & 2.5 & 6.9 \\
        \bottomrule
    \end{tabular}
\end{table}

\section{Discussion}
\label{sec:discussion}

\paragraph{Why does LoRA match FFT despite high intrinsic rank?}
The SVD analysis (Section~\ref{sec:intrinsic_rank_results}, Fig.~\ref{fig:cumulative_energy}) reveals that FFT weight deltas are high-dimensional, yet LoRA at $r{=}32$ matches FFT task performance (Table~\ref{tab:atp_overall}). This suggests that the task-critical adaptation occupies a smaller subspace than the full weight update for this experimental setup, and that the remaining spectral energy may represent noise that does not contribute to downstream manipulation performance. The saturation curve in Fig.~\ref{fig:rank_scaling} further supports this interpretation: performance plateaus well before LoRA can capture the majority of the spectral energy. Unlike OpenVLA, which tested only two LoRA configurations, the full rank sweep presented here traces the complete saturation curve and demonstrates that the plateau is robust across ranks in this configuration.
\paragraph{Architectural roles shape adaptation requirements.}
Three findings from Section~\ref{sec:results} converge on a coherent picture of how $\pi_0$'s architecture shapes adaptation. 
First, per-layer rank heterogeneity (Fig.~\ref{fig:intrinsic_rank_per_layer}) reveals that intrinsic dimensionality varies substantially across layers. The VLM's MLP final-layer spike is consistent with a representational mismatch originating from PaliGemma's pre-training, where this layer was optimized for next-token language prediction via cross-entropy loss. Although $\pi_0$'s pre-training on robot data with a flow-matching objective~\cite{black2024pi0} partially restructures these representations, the last layer evidently requires further high-rank adaptation when transferring to a new embodiment. The action expert's last layer, initialized from scratch for action generation, shows the opposite trend. Despite this heterogeneity, uniform LoRA allocation performs on par with asymmetric strategies (Section~\ref{sec:results}), suggesting that over-provisioning some layers while under-provisioning others is tolerable in practice.
Second, SigLIP's high normalized intrinsic rank (Section~\ref{sec:intrinsic_rank_results}) offers an explanation why LoRA is insufficient for this component (Section~\ref{sec:siglip_results}): the visual domain shift from pre-training to the target workspace requires high-rank adaptation that low-rank updates cannot provide. Because the main LoRA comparison leaves SigLIP unconstrained, only the lower-rank VLM and action expert are limited by the adapter budget, which likely contributes to the robustness of LoRA performance across ranks.
Third, the VLM freezing result (Section~\ref{sec:results}) combined with SigLIP's critical role indicates that embodiment adaptation is not exclusively a motor control problem---it requires both semantic and visual adaptation of the VLM to the target environment.
\paragraph{Practical implications.}
The results suggest a concrete practitioner recipe: apply LoRA at $r{=}16$--$32$ with uniform allocation to the VLM and action expert, while fully fine-tuning SigLIP. This configuration shows no statistically significant performance drop compared to FFT while reducing static peak VRAM from 36.2\,GiB to 10.8\,GiB (Table~\ref{tab:memory}), fitting within a single consumer GPU. A fully LoRA-based pipeline---including SigLIP---further reduces trainable parameters to 2.7\,\% of the full model, but incurs a substantial performance penalty (Table~\ref{tab:atp_overall}).
\paragraph{Limitations.}
Our study is conducted on a single robotic platform with a single VLA architecture. While the tasks span a range of manipulation challenges, generalization to other embodiments or VLA architectures requires further investigation. The evaluation uses 20 rollouts per task, which provides reasonable statistical power for ATP but limits the detection of small effects.

A further limitation concerns the LoRA scaling convention. We fix $\alpha{=}r$ across the full rank sweep, following the standard protocol of Hu et al.~\cite{hu2021lora}. Kalajdzievski~\cite{kalajdzievski2023rslora} shows that the resulting $\alpha/r$ scaling can induce gradient collapse at large $r$, slowing learning rather than reflecting a true capacity ceiling. The plateau beyond $r{=}32$ (Finding 2) may therefore partly reflect this scaling rule. Evaluating rank-stabilized scalings such as rsLoRA ($\alpha/\sqrt{r}$)~\cite{kalajdzievski2023rslora} is a clear direction for future work.

\section{Conclusion}
\label{sec:conclusion}

We presented a systematic study of LoRA-based fine-tuning for $\pi_0$, a flow-matching VLA, on four precision industrial manipulation tasks, on one physical setup. Our evaluation across six LoRA ranks, two asymmetric allocation strategies, two component-freezing ablations, and an intrinsic rank analysis of FFT weight updates yields three findings. 

First, LoRA shows no statistically significant performance drop: most tested configurations show matching performance for a setup with LoRA for VLM and AE, together with full SigLIP fine-tuning compared to FFT. For this setup, performance saturates at $r\sim32$, with higher ranks providing no additional benefit while increasing memory consumption. Second, concentrating capacity on either component yields no measurable benefit, and VLM adaptation is essential for task success. Third, vision encoder adaptation via FFT seems essential for LoRA's strong performance: freezing SigLIP or applying LoRA to it significantly degrades task progress for the tested setup.

Furthermore, SVD analysis suggests that although FFT MLP weight deltas are high-dimensional, the task-critical adaptation resides in a low-rank subspace, supporting the distributed adaptation observed across the model components. These results suggest that LoRA with uniform allocation at $r{=}32$ on the VLM and action expert, combined with full fine-tuning of SigLIP, provides a practical and resource-efficient strategy for deploying flow-matching VLAs on industrial hardware. In this case static VRAM requirements for parameters and optimizer states, excluding activation memory, are reduced by up to 70\,\%  relative to FFT.


\begin{credits}
\subsubsection{\discintname}
The authors have no competing interests to declare that are relevant to the content of this article.

\subsubsection{\ackname} This work was partially funded by the Fraunhofer Research Programs under Grant No. Attract 40-11894.

\end{credits}

\bibliographystyle{splncs04}
\bibliography{refs}

\end{document}